\documentclass{article}
\usepackage[final]{nips_2018}
\usepackage{times}
\usepackage{latexsym}
\usepackage{graphicx}
\usepackage{subfigure}
\usepackage{booktabs} 
\usepackage{hyperref}
\usepackage{latexsym}
\usepackage{amsfonts,amssymb}
\usepackage{amsmath}
\usepackage{graphicx}
\usepackage{algorithm}
\usepackage{algorithmic}
\usepackage{float}
\usepackage{array}
\usepackage{multirow}
\usepackage{url}
\usepackage{makecell}

\title{Hungarian Layer: A Novel Neural Layer for Paraphrase Identification}
\author{Han Xiao, Yidong Chen, Xiaodong Shi \\
	Xiamen Univeristy \\
	bookman@xmu.edu.cn
}
\begin{document}
\newcommand\BibTeX{B{\sc ib}\TeX}
\maketitle

\begin{abstract}
Paraphrase identification is an important topic in artificial intelligence and this task is often tackled as sequence alignment and matching. Traditional alignment methods take advantage of attention mechanism, which is a soft-max weighting technique. Weighting technique could pick out the most similar/dissimilar parts, but is weak in modeling the aligned unmatched parts, which are the crucial evidence to identify paraphrase. In this paper, we empower neural architecture with Hungarian algorithm to extract the aligned unmatched parts. Specifically, first, our model applies BiLSTM to parse the input sentences into hidden representations. Then, Hungarian layer leverages the hidden representations to extract the aligned unmatched parts. Last, we apply cosine similarity to metric the aligned unmatched parts for a final discrimination. Extensive experiments show that our model outperforms other baselines, substantially and significantly.
\end{abstract}

\section{Introduction}
Paraphrase identification is an important topic in artificial intelligence and this task justifies whether two sentences expressed in various forms are semantically similar, \cite{Chitra2016Plagiarism}. For example, \textit{``On Sunday, the boy runs in the yard''} and \textit{``The child runs outside at the weekend''} are identified as paraphrase. This task directly benefits many industrial applications, such as plagiarism identification \cite{Chitra2016Plagiarism}, machine translation \cite{Kravchenko2017Paraphrase} and removing redundancy questions in Quora website \cite{Wang2017Bilateral}. Recently, there emerge many methods, such as ABCNN \cite{Yin2015ABCNN}, Siamese LSTM \cite{Wang2017Bilateral} and L.D.C \cite{Wang2016Sentence}.

Conventionally, neural methodology aligns the sentence pair and then generates a matching score for paraphrase identification, \cite{Wang2016Sentence, Wang2017Bilateral}. Regarding the alignment, we conjecture that \textbf{\textit{the aligned unmatched parts are semantically critical}}, where we define the corresponded word pairs with low similarity as aligned unmatched parts. For an example: \textit{``On Sunday, the boy runs in the yard''} and \textit{``The child runs inside at the weekend''}, the matched parts (\textit{i.e. (Sunday, weekend), (boy, child), run}) barely make contribution to the semantic sentence similarity, but the unmatched parts \textit{(i.e. ``yard'' and ``inside'')} determine these two sentences are semantically dissimilar. For another example: \textit{``On Sunday, the boy runs in the yard''} and \textit{``The child runs outside at the weekend''}, the aligned unmatched parts \textit{(i.e. ``yard'' and ``outside'')} are semantically similar, which makes the two sentences paraphrase. \textit{\textbf{In conclusion, if the aligned unmatched parts are semantically consistent, the two sentences are paraphrase, otherwise they are non-paraphrase.}}

\begin{figure}
	\centering
	\includegraphics[width=0.75\linewidth]{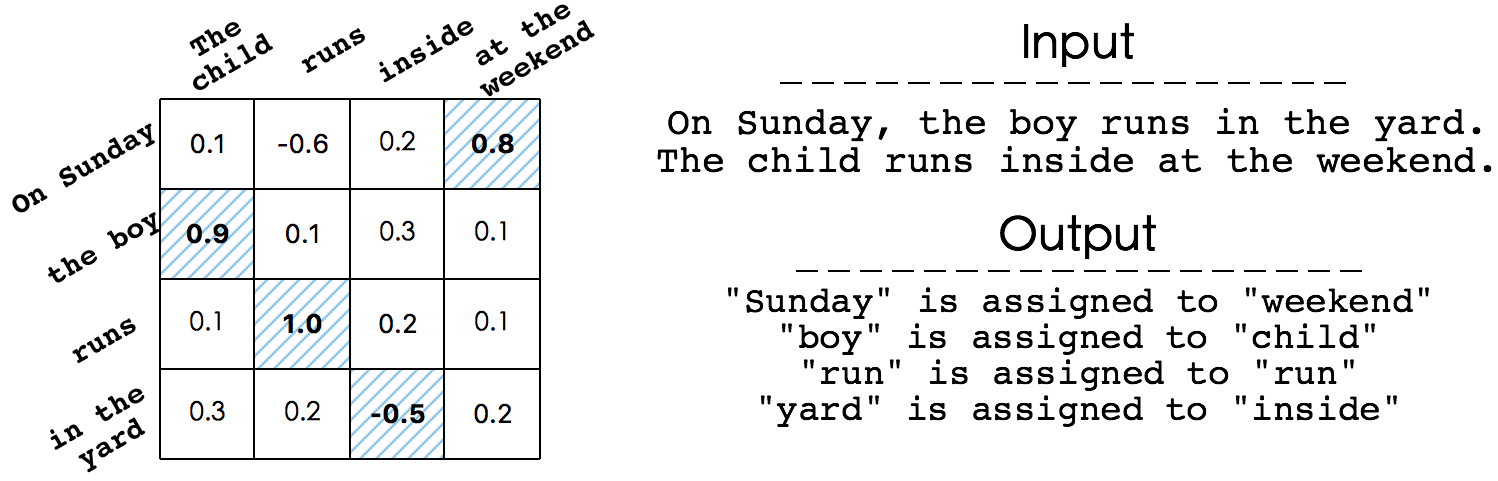}
	\caption{Sentence Matching Formulated as Task Assignment. ``Sunday'', ``The boy'', ``The child'', ``run'' and ``weekend'' are assigned to the corresponding parts, exclusively. Thus, as the aligned unmatched part, ``yard'' is assigned to ``inside'', which is the evidence of non-paraphrase.}
	\label{fig:sentencematching}
\end{figure}

Traditional alignment methods take advantage of attention mechanism \cite{Wang2017Bilateral}, which is a soft-max weighting technique. Weighting technique could pick out the most similar/dissimilar parts, but is weak in modeling the aligned unmatched parts, which are the crucial evidence to identify paraphrase. For the input sentences in Figure \ref{fig:sentencematching}, the weight between ``Sunday'' and ``run'' is lower than the weight between ``yard'' and ``inside'', but the former weight is not the evidence of paraphrase/non-paraphrase, because the former two words that are most dissimilar should not be aligned for an inappropriate comparison.

To extract the aligned unmatched parts, in this paper, we embed Hungarian algorithm \cite{Wright1990Speeding} into neural architecture as Hungarian layer (Algorithm \ref{alg}).  Illustrated in Figure \ref{fig:sentencematching}, the alignment in sentence matching could be formulated as the task-assignment problem, which is tackled by Hungarian algorithm. Simply, Hungarian algorithm works out the theoretically optimal alignment relationship in an exclusive manner and  \textit{\textbf{the exclusiveness characterizes the aligned unmatched parts.}} For the example in Figure \ref{fig:sentencematching}, because Hungarian layer allocates the aligned pairs with exclusiveness, the matched parts (i.e \textit{(Sunday, weekend), (boy, child), run}) are aligned firstly, then the word \textit{``yard''} would be assigned to the word \textit{``inside''} with a negative similarity, making a strong evidence for discrimination.

Specifically, our model performs this task in three steps. First, our model applies BiLSTM to parse the input sentences into hidden representations. Then, Hungarian layer leverages the hidden representations to extract the aligned unmatched parts. Last, we apply cosine similarity to metric the aligned unmatched parts for a final discrimination. Regarding the training process of Hungarian layer, we modify the back-propagation algorithm in both directions. In the forward pass, Hungarian layer works out the alignment relationship, according to which, the computational graph is dynamically constructed, as demonstrated in Figure \ref{fig:dynamics}. Once the computational graph has been dynamically constructed, the backward propagation could be performed as usual in a conventional graph.

We conduct our experiments on the public benchmark dataset of ``Quora Question Pairs'' for the task of paraphrase identification. Experimental results demonstrate that our model outperforms other baselines extensively and significantly, which verifies our theory about the aligned unmatched parts and illustrates the effectiveness of our methodology.

\textbf{Contributions.} \textbf{(1.)} We offer a new perspective for paraphrase identification, which focuses on the aligned unmatched parts of two sentences. Accordingly, we propose the Hungarian layer to extract the aligned unmatched parts. The proposed method can achieve hard and exclusive alignments between two sequences, while we can learn parameters by end-to-end back-propagation. \textbf{(2.)} Our model outperforms other baselines extensively, verifying the effectiveness of our theory and method.

\textbf{Organization.} In Section 2, we survey the related work of paraphrase identification and dynamic differentiable computational graphs. In Section 3, we introduce our neural architecture. In Section 4, we conduct the experiments. In Section 5, we conclude our paper and publish our codes.

\section{Related Work}
We have surveyed this task and categorized related papers into three lines.
  
\subsection{Non-Neural Architecture for Paraphrase Identification}
The topic of paraphrase identification raises in the last decade. The development has been through four stages before neural architectures: \textit{word specific, syntactic tree specific, semantic matching and probabilistic graph modeling.}

Firstly, \cite{Bilotti2007Structured} focuses on simple surface-form matching between bag-of-words, which produces poor accuracy, because of word ambiguities and syntactic complexity. Therefore, syntactic analysis is introduced into this task for semantic understanding, such as deeper semantic analysis \cite{Shen2007Using}, quasi-synchronous grammars \cite{Wang2009What} and tree edit distance \cite{Heilman2010Tree}. Notably, most of these methods compare the grammar tree (\textit{e.g. syntactic tree, dependency tree, etc.}) of sentence pair. Further, semantic information such as negation, hypernym, synonym and antonym is integrated into this task for a better prediction precision, \cite{Lai2014Illinois}. Finally, \cite{Yao2013Semi} leverages a semi-Markov CRF to align phrases rather than words, which consumes too many resources for industrial applications.

In summary, the advantage of this branch, which roots the foundation in linguistics, is semantically interpretable, while the disadvantage is too simple to understand complex language phenomenon. 

\subsection{Neural Architecture for Paraphrase Identification: Independent Sentence Encoder}
With the popularity of deep neural network, some neural architectures are proposed to analyze the complex language phenomenon in a data-fitting way, which promotes the performance. First of all, the neural network extracts the abstracted features from each sentence independently, then measures the similarity of the abstracted feature pair. There list two frameworks: \textit{CNN-based and RAE-based.}

Commonly, CNN could be treated as n-gram method, which corresponds to language model. Specifically, \cite{yu2014deep} applies a bi-gram CNN to jointly model source and target sequences. \cite{Yang2015WikiQA} achieves a better performance by following this work. \cite{socher2011dynamic} has proposed a RAE based model to characterize phrase-level representation, which promotes simple pooling method, \cite{Blacoe2012A}. Multi-perspective methods \cite{Wang2017Bilateral} take the advantage of multiple metric aspects to boost the accuracy.

In summary, the advantage of this branch is to model complex and ambiguous linguistic phenomenon in a black-box style. However, the disadvantage is that the encoder could not adjust the abstracted representations according to the correlation of sentence pair, making an imperfect matching process.

\subsection{Neural Architecture for Paraphrase Identification: Interdependent Sentence Encoder}
To emphasize the correlation of sentence pair in encoder, the researchers propose the attention-based neural architectures, which guide the encoding process according to the corresponding part. There introduce the representative methods: ABCNN \cite{Yin2015ABCNN} and L.D.C \cite{Wang2017Bilateral}.

ABCNN is a CNN-based model. In a single stage, this model computes the attention similarity matrix for the convolution layer, then sums out each row and column as the weighs of pooling layer. The output of convolution layer is weighted by pooling layer in an average manner as the output of this stage. ABCNN could stack at most three stages. This method achieves satisfactory performance in many tasks, because of modeling correlation in sentence encoder. L.D.C model \cite{Wang2016Sentence} is an attention-based method, which decomposes the hidden representations into similar and dissimilar parts, then respectively processes each parts to generate the final result. Notably, L.D.C is the state-of-the-art method.

In summary, the advantage of this branch is to model alignment or correlation in the encoding process. However, the disadvantage is to focus on the matched parts, rather than the unmatched parts, which are critical in this task as previously discussed.

\subsection{Dynamic Differentiable Computational Graphs}
Neural Turing Machine (NTM) \cite{Graves2014Neural, Gulcehre2016Dynamic} is a seminal work to implement instrument-based algorithm in the neural architecture, which attempts to express algorithms by simulating memory and controller. However, NTM leverages the weighting technique, which involves too much noise and makes the learned algorithm fuzzy. Thus, we propose a hard way to embed algorithms into neural architectures.

There also exist some papers for dynamical computational graph construction. At the lower level, pointer-switch networks \cite{Gulcehre2016Pointing} are a kind of dynamic differentiable neural model. At the higher level, some architecture search models \cite{Pham2018Efficient, Fernando2017PathNet} construct new differentiable computational graphs dynamically at every iteration.

\begin{figure}
	\centering
	\includegraphics[width=\linewidth]{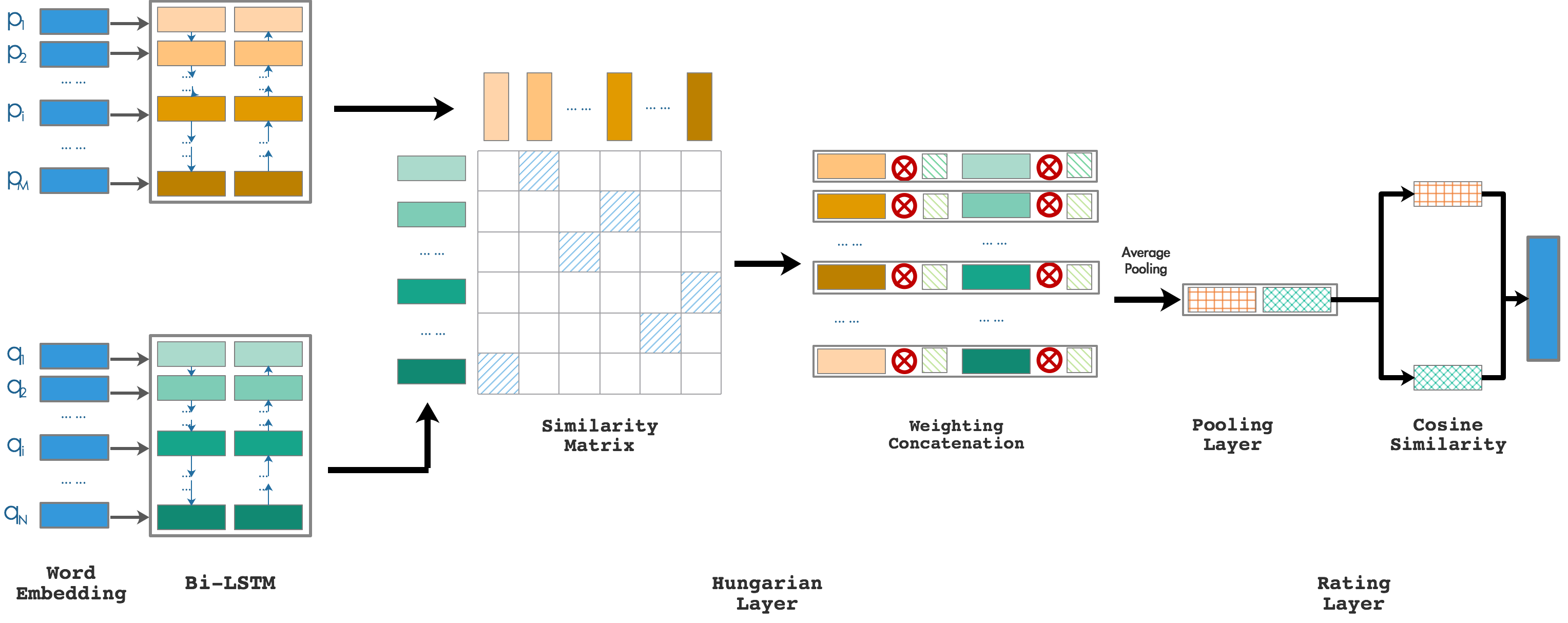}
	\caption{Our Neural Architecture. The sentence pair corresponds to word sequences $(p_1, p_2, ..., p_M)$ and $(q_1, q_2, ..., q_N)$. First, word embeddings are parsed by BiLSTM to generate the hidden representations. Then, the hidden representations of BiLSTM are processed by Hungarian layer, the outputs of which correspond to the weighting concatenation of aligned representations. Last, the results of Hungarian layer are measured by cosine similarity.}
	\label{fig:methodology}
\end{figure}

\section{Methodology}
First, we introduce the basic components of our neural architecture. Then, we analyze the training process of Hungarian layer, that how to dynamically construct the computational graph.

\subsection{Neural Architecture}
Our neural architecture is illustrated in Figure \ref{fig:methodology}. Basically our model is composed by four components, namely, word embedding, bi-directional LSTM (BiLSTM), Hungarian layer and cosine similarity.

\textbf{Word Embedding.} The goal of this layer is to represent each word $w_{s,i}$ in every sentence $s$ with $d$-dimensional semantic vectors. The word representations, which are pre-trained by GloVe \cite{Pennington2014Glove}, are unmodified within the learning procedure. The inputs of this layer are a pair of sentences as word sequences $p=(p_1, p_2, ..., p_M)$ and $q=(q_1, q_2, ..., q_N)$, while the outputs are corresponding embedding matrices as $\mathbf{P=(p_1, p_2, ..., p_M)}$ and $\mathbf{Q=(q_1, q_2, ..., q_N)}$.

\textbf{Bi-Directional LSTM (BiLSTM).} The purpose of this layer is to transform lexical representations to hidden contextual representations. For hidden contextual encoding, we employ a parameter-shared bi-directional LSTM (BiLSTM) \cite{Hochreiter1997Long} to parse the word embeddings into hidden representations, mathematically as:
\begin{eqnarray}
	\mathbf{h_{i}^f} = \mathbf{LSTM^{forward}}(\mathbf{h_{i-1}^f, s_i}) \\
	\mathbf{h_{i}^b} = \mathbf{LSTM^{backward}}(\mathbf{h_{i+1}^b, s_i}) 
\end{eqnarray}
where $\mathbf{h_i} = \mathbf{[h^f_i, h^b_i]}$ is the $i$-th hidden representation and $\mathbf{s_i}$ corresponds to the $i$-th word embedding in the source/target sentence or $\mathbf{p_i}$/$\mathbf{q_i}$.

\textbf{Hungarian Layer.} This layer, which is the matching component of our model, extracts the aligned unmatched parts from the source and target sentences. This layer is composed by two sequential stages.

Algorithm \ref{alg} demonstrates the first stage. The objective of this stage is to align the source and target hidden representations. The inputs of this stage are $M$ source hidden representation vectors $\{\mathbf{s_i}\}$ and $N$ target hidden representation vectors $\{\mathbf{t_i}\}$, while the outputs of this stage are $M$ aligned hidden representation vector pairs $(\mathbf{a_i, b_i)}$, assuming $M \le N$, where $\mathbf{a_i/b_i}$ corresponds to the $i$-th aligned source/target hidden representation vector, respectively.

Specifically in this stage, there are totally three steps. First, the input hidden representations are crossly dotted to generate the pairwise similarity matrix $w_{i,j}$. Then, Hungarian algorithm works out the aligned source-target position pairs $(g_i, h_i)$ with this similarity matrix. For example in Figure \ref{fig:sentencematching}, assuming the left/top sentence indicates the source/target sequence, the aligned source-target position pairs are listed as $\{(3,2), (2,1), (1,4), (4,3)\}$. Last, the input hidden representation vectors $\mathbf{\{s_i\}, \{t_j\}}$ are re-organized into the aligned source-target hidden representation vector pairs $\mathbf{(a_i=s}_{g_i}, \mathbf{b_i=t}_{h_i})$, according to the aligned source-target position pairs $(g_i, h_i)$.

The second stage attempts to extract the aligned unmatched parts by weighting the aligned hidden representations $\mathbf{(a_i, b_i)}$ from the first stage. Required by extracting the unmatched parts, if two aligned representations are matched, the weight for them should be small, otherwise, large dissimilarity leads to large weight. For this reason, we introduce cosine dissimilarity, mathematically as: 
\begin{eqnarray}
	\alpha_i = 1 - m_i
\end{eqnarray}
where $\alpha_i$ is the $i$-th aligned cosine dissimilarity and $m_i$ is the $i$-th aligned cosine similarity from the first stage. Thus, the aligned hidden representations are concatenated and then weighted by cosine dissimilarity:
\begin{eqnarray}
	\mathbf{R_i} = \alpha_i \otimes \mathbf{[a_i, b_i]}
\end{eqnarray}
where $\mathbf{R_i}$ is the $i$-th output of Hungarian layer, $\mathbf{a_i/b_i}$ is the $i$-th aligned source/target hidden representation generated by Algorithm \ref{alg} and $\otimes$ is the scalar-vector multiplication. \textit{Actually in the practical setting, most of cosine dissimilarity approach $0$ and the remaining hidden representations indicate the aligned unmatched parts.} 

\begin{algorithm}[t]
	\renewcommand{\algorithmicrequire}{\textbf{Input:}}
	\renewcommand{\algorithmicensure}{\textbf{Output:}}
	\caption{Hungarian Layer: First Stage}
	\label{alg}
	\begin{algorithmic}[1]
			\REQUIRE Source and target sentence hidden representations: $\{\mathbf{s_i}\}$ and $\{\mathbf{t_i}\}$.
			\ENSURE $\{(\mathbf{a_i, b_i}, m_i)\}$, where $\mathbf{a_i}$ and $\mathbf{b_i}$ mean the $i$-th aligned hidden representations for source and target respectively, and $m_i$ means the corresponding similarity.
			\STATE Generate the pairwise similarity matrix: $$w_{i,j} = \frac{dot(\mathbf{s_i, t_j})}{\mathbf{|s_i||t_j|}}$$ where $dot$ is the dot product and $|\cdot|$ is the length of vector.
			\STATE Perform Hungarian algorithm \cite{Wright1990Speeding} to assign the aligned position pairs $\{(g_i, h_i)\}$, where $g_i/h_i$ is $i$-th aligned source/target position of the sentence pair.
			\FORALL{$i \in [1...M]$, where $M$ is the length of source sentence.}
				\STATE Compute $m_i = w_{g_i, h_i}$, where $m_i$ is the pairwise similarity for $i$-th matched position.
			\ENDFOR
			\STATE \textbf{return} $\{(\mathbf{a_i} = \mathbf{s}_{g_i}, \mathbf{b_i} = \mathbf{t}_{h_i}, m_i)\}$, where $\mathbf{a_i/b_i}$ corresponds to the $i$-th aligned source/target hidden representation, while $(g_i, h_i)$ is the $i$-th aligned source-target position pair,  $\mathbf{\{s_i/t_i\}}$ are the input source/target hidden representation vectors and $m_i$ is the $i$-th aligned cosine similarity.
	\end{algorithmic}
\end{algorithm}

\textbf{Cosine Similarity.} Last, we average the concatenated hidden representations as the final sentence representation $\mathbf{R} = \frac{1}{M} \sum_{i=1}^{M} \mathbf{R_i}$, which is a conventional procedure in neural natural language processing, \cite{Wang2016Sentence}. Then, we employ a cosine similarity as the output:
\begin{eqnarray}
	y = \mathbf{\frac{R_p^TR_q}{|R_p||R_q|}}
\end{eqnarray}
where $y$ is the matching score, $|\cdot|$ is the length of vector and $\mathbf{R_p}$/$\mathbf{R_q}$ is the corresponding source/target part of the final sentence representation $\mathbf{R}$. Thus, our output ranges in $[-1, +1]$, where $+1$ means the two sentences are similar/paraphrase, and $-1$ means otherwise. For further evaluation of accuracy, we also apply a threshold learned in the development dataset to binary the cosine similarity as paraphrase/non-paraphrase. Notably, the introduction of concatenation layer facilitates the inference and training of Hungarian layer.

\subsection{Training Hungarian Layer}
\begin{figure}
	\centering
	\includegraphics[width=0.65\linewidth]{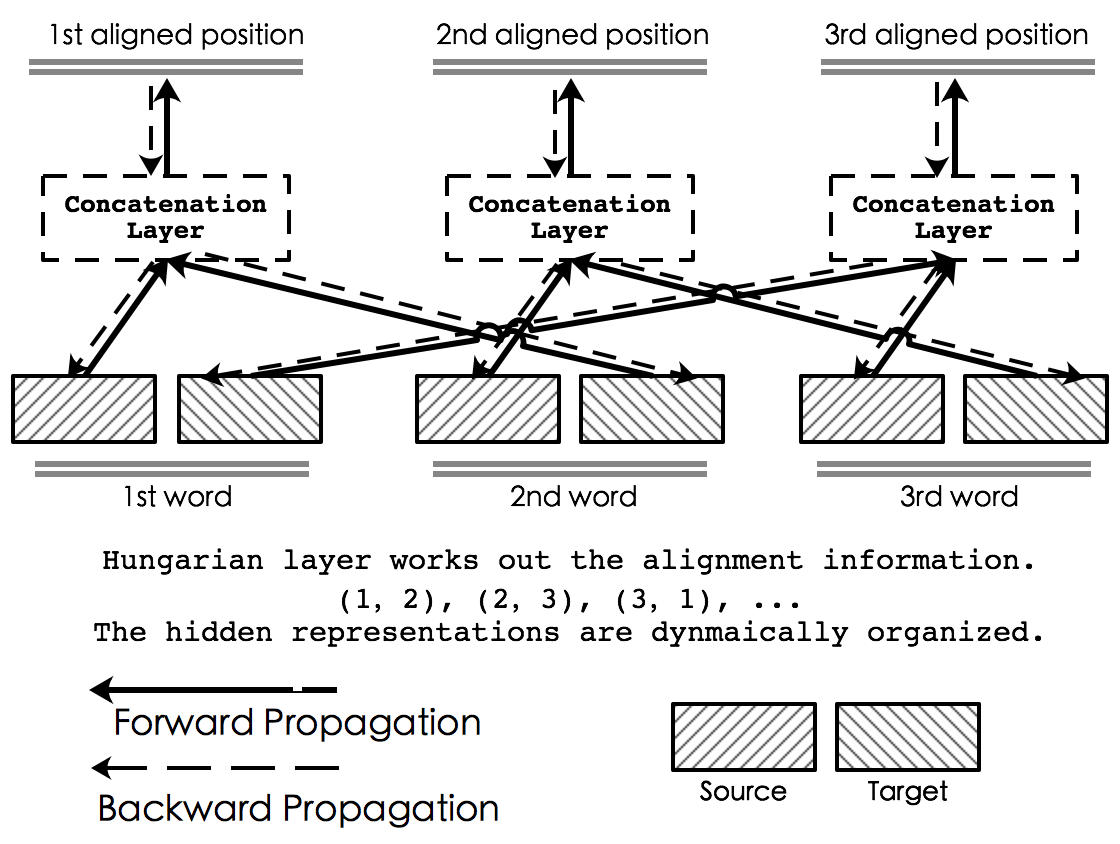}
	\caption{Illustration of Dynamically Constructed Computational Graph. The solid lines correspond to the forward propagation, while the dashed lines correspond to the backward propagation. Hungarian layer computes the aligned position pairs, according to which, the hidden representations are dynamically connected to the next layer. For example, $(1, 2)$ means 1st word in source corresponds to 2nd word in target, while the 1st source and 2nd target word representations are jointly linked to the 1st aligned position of concatenation layer in the forward propagation. Backward propagation is performed through the dynamically constructed links between layers without any branching and non-differentiated issues.}
	\label{fig:dynamics}
\end{figure}

Previously discussed, Hungarian algorithm is embedded into neural architecture, making a challenge for learning process. We tackle this issue by modifying the back-propagation algorithm in a dynamically graph-constructing manner. In the forward pass, we dynamically construct the links between Hungarian layer and the next layer, according to the aligned position pairs, while in the backward process, the back-propagation is performed through the dynamically constructed links. Next, we illustratively exemplify how the computational graph is dynamically constructed in Hungarian layer as Figure \ref{fig:dynamics} shows.

As Figure \ref{fig:dynamics} shows, in the forward propagation, Hungarian algorithm works  out the aligned position pairs, according to which, neural components are dynamically connected to the next layer. For the example of Figure \ref{fig:dynamics}, the 1st source and 2nd target word representations are jointly linked to the 1st aligned position of concatenation layer. \textbf{\textit{Once the computational graph has been dynamically constructed in the forward pass, the backward process could propagate through the dynamically constructed links between layers, without any branching and non-differentiated issues.}} For the example in Figure \ref{fig:dynamics}, the backward pass firstly propagates to the 1st aligned position of concatenation layer, then respectively propagates to 1st source and 2nd target word representations. In this way, the optimization framework could still adjust the parameters of neural architectures in an end-to-end manner.

\section{Experiment}
In this section, we verify our model performance on the famous public benchmark dataset of ``Quora Question Pairs''. First, we introduce the experimental settings, in Section 4.1. Then, in Section 4.2, we conduct the performance evaluation. Last, in order to further test our assumptions, that the aligned unmatched parts are semantically critical, we conduct a case study for illustration in Section 4.3.

\subsection{Experimental Setting}
We initialize the word embedding with 300-dimensional GloVe \cite{Pennington2014Glove} word vectors pre-trained in the 840B Common Crawl corpus \cite{Pennington2014Glove}. For the out-of-vocabulary (OOV) words, we directly apply zero vector as word representation. Regarding the hyper-parameters, we set the hidden dimension as 150 for each BiLSTM. To train the model, we leverage AdaDelta \cite{Zeiler2012ADADELTA} as our optimizer, with hyper-parameters as moment factor $\eta=0.6$ and $\epsilon=1 \times 10^{-6}$. We train the model until convergence, but at most 30 rounds. We apply the batch size as $512$.

\subsection{Performance Evaluation} 

\textbf{Dataset.} Actually, to demonstrate the effectiveness of our model, we perform our experiments on the famous public benchmark dataset of ``Quora Question Pairs'' \footnote{The url of the dataset: \url{https://data.quora.com}}. For a fair comparison, we follow the splitting rules of \cite{Wang2017Bilateral}. Specifically, there are over 400,000 question pairs in this dataset, and each question pair is annotated with a binary value indicating whether the two questions are paraphrase of each other or not. We randomly select 5,000 paraphrases and 5,000 non-paraphrases as the development set, and sample another 5,000 paraphrases and 5,000 non-paraphrases as the test set. We keep the remaining instances as the training set. 

\textbf{Baselines.} To make a sufficient comparison, we choose five state-of-the-art baselines: Siamese CNN, Multi-Perspective CNN, Siamese LSTM, Multi-Perspective LSTM, and L.D.C. Specifically, Siamese CNN and LSTM encode the two input sentences into two sentence vectors by CNN and LSTM, respectively, \cite{Wang2016Semi}. Based on the two sentence vectors, a cosine similarity is leveraged to make the final decision. Multi-Perspective methods leverage different metric aspects to promote the performance, \cite{Wang2017Bilateral}. L.D.C model \cite{Wang2016Sentence} is an attention-based method, which decomposes the hidden representations into similar and dissimilar parts. L.D.C is a powerful model which achieves the state-of-the-art performance. 

We have tested L.D.C. and our model five times to evaluate the mean and variance, then perform the test for statistical significance.

\begin{table*}
	\caption{Performance Evaluation on ``Quora Question Pairs''.}
	\centering
	\label{tab:paraphrase}
	\renewcommand\arraystretch{1.1}
	\begin{tabular}{c c}
		\Xhline {1.2pt} Methods & Accuracy (\%)\\
		\Xhline {1.2pt}
		Siamese CNN \cite{Wang2016Semi} & 79.60 \\
		Multi-Perspective CNN \cite{Wang2017Bilateral} & 81.38 \\
		Siamese LSTM \cite{Wang2016Semi} & 82.58 \\
		Multi-Perspective LSTM \cite{Wang2017Bilateral} & 83.21 \\
		L.D.C. \cite{Wang2016Sentence} & 84.75 $\pm$ 0.42 \\
		\hline
		\hline 
		Our Model & \textbf{85.53 $\pm$ 0.18 $^*$} \\
		\Xhline {1.2pt}
	\end{tabular}
\end{table*}

\renewcommand{\thefootnote}{}
\footnotetext{$^*$ We apply t-test and $p=0.003 < 0.01$. Thus, the improvement is statistically significant.}

\begin{figure*}[t]
	\centering
	\includegraphics[width=0.90\linewidth]{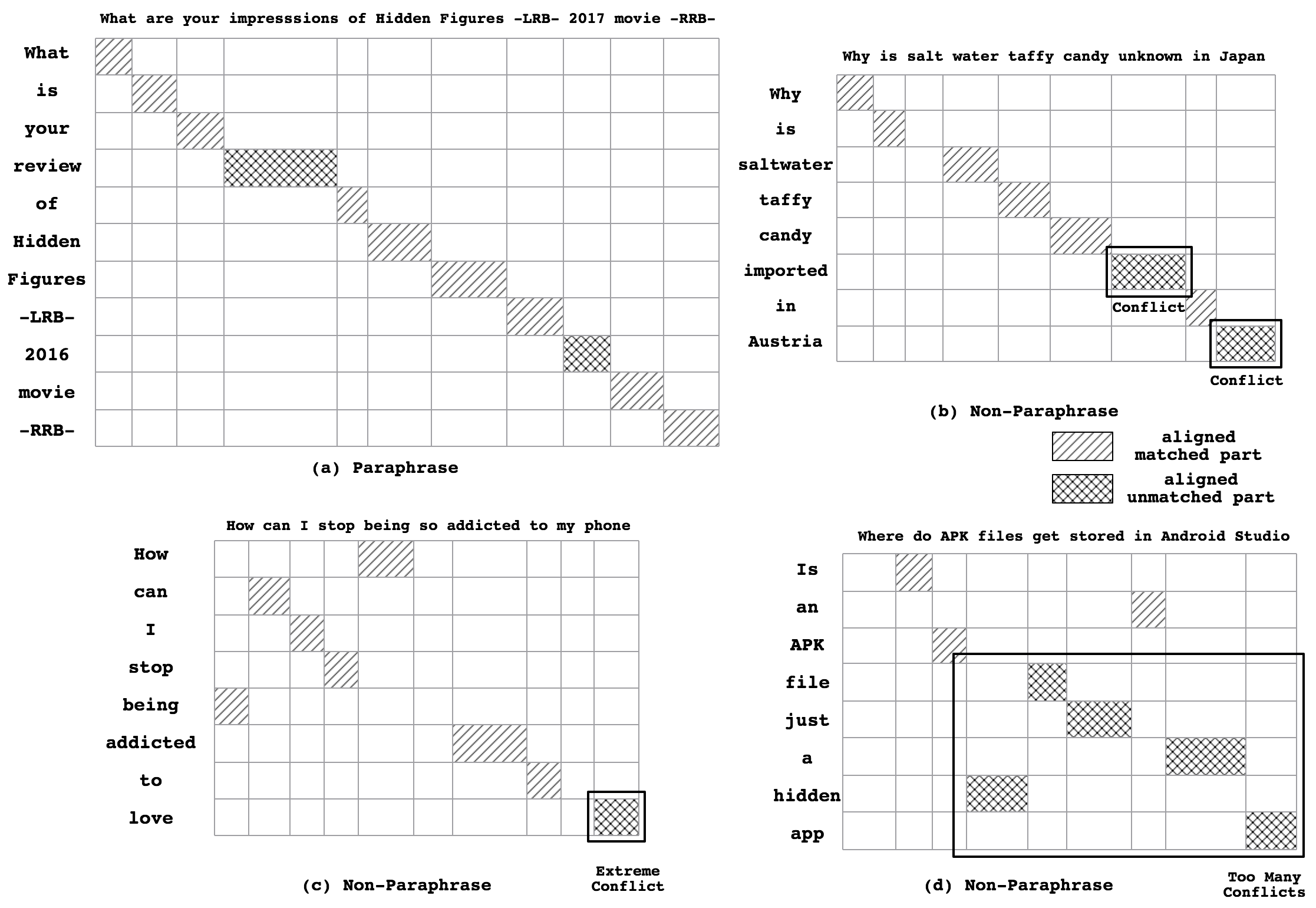}
	\caption{Illustration of Case Study. The slashed grids correspond to the aligned matched parts, while the crossed ones indicate the aligned unmatched parts.}
	\label{fig:h-layer}
\end{figure*}

\textbf{Results.} Our results are reported in Table \ref{tab:paraphrase}. We can conclude that:
\begin{enumerate}
	\item Our method outperforms all the baselines, which illustrates the effectiveness of our model.
	\item In order to evaluate the reliability of the comparison between L.D.C and our model, the results are tested for statistical significance using t-test. In this case, we obtain a p-value = 0.003 $<$ 0.01. Therefore, the null hypothesis that values are drawn from the same population (i.e., the accuracies of two approaches are virtually equivalent) can be rejected, which means that the improvement is statistically significant.
	\item Compared with Siamese LSTM \cite{Wang2016Semi}, which lacks the matching layer, our model could precisely align the input sentences. Thus, our method promotes the performance.
	\item Compared with L.D.C. \cite{Wang2016Sentence}, which is an attention-based method and still analyzes the dissimilar part, our model could exactly extract the aligned unmatched parts rather than the fuzzy dissimilar parts. Thus, our performance is better.
	\item Notably, L.D.C. is a very complex model, which is beaten by our simple model within a statistically significant improvement. This comparison illustrates our model is indeed simple but effective. Thus it is very suitable for industrial applications.  
\end{enumerate}

\subsection{Case Study}
We have conducted a case study in the practical setting of ``Quora Question Pairs'' with our model for paraphrase identification. Illustrated in Figure \ref{fig:h-layer}, the slashed grids correspond to the aligned matched parts, while the crossed ones indicate the aligned unmatched parts. Notably, we mark the pairwise similarity below $0.3$ as unmatched in this case study.

For the example of (a), there exist two input sentences: ``What is your review of Hidden Figures -LRB- 2016 movie -RRB-'' and ``What are your impressions of Hidden Figures -LRB- 2017 movie -RRB-''. From our case analysis, most of the aligned parts are matched, while minor aligned unmatched parts are similar. Thus, our method justifies the two sentences as paraphrase. This is accorded to our assumption.

For the example of (b), there exist two input sentences: ``Why is saltwater taffy candy imported in Austria'' and ``Why is salt water taffy candy unknown in Japan''. There are two unmatched parts that ``imported/unknown'' and ``Austria/Japan'', which are conflicted. Thus, the case is classified as non-paraphrase.

For the example of (c), the two sentences are: ``How can I stop being addicted to love'' and ``How can I stop being so addicted to my phone''. From our case analysis, there is an extreme conflict that ``love/phone'', making this case non-paraphrase, according to our assumption.

For the example of (d), the two sentences are: ``Is a(n) APK file just a hidden app'' and ``Where do APK files get stored in Android Studio''. As we know, there are too many conflicts in this case, making a very dissimilar score as non-paraphrase.

In summary, this case study justifies our assumption that ``the aligned unmatched parts are semantically critical''.

\section{Conclusion}
In this paper, we leverage Hungarian algorithm to design Hungarian layer, which extracts the aligned matched and unmatched parts exclusively from the sentence pair. Then our model is designed by assuming the aligned unmatched parts are semantically critical. Experimental results on benchmark datasets verify our theory and demonstrate the effectiveness of our proposed method.

\newpage

\bibliography{Ref}

\begin{thebibliography}{25}
\providecommand{\natexlab}[1]{#1}
\providecommand{\url}[1]{\texttt{#1}}
\expandafter\ifx\csname urlstyle\endcsname\relax
  \providecommand{\doi}[1]{doi: #1}\else
  \providecommand{\doi}{doi: \begingroup \urlstyle{rm}\Url}\fi

\bibitem[Bilotti et~al.(2007)Bilotti, Ogilvie, Callan, and
  Nyberg]{Bilotti2007Structured}
Bilotti, Matthew~W., Ogilvie, Paul, Callan, Jamie, and Nyberg, Eric.
\newblock Structured retrieval for question answering.
\newblock In \emph{International ACM SIGIR Conference on Research and
  Development in Information Retrieval}, pp.\  351--358, 2007.

\bibitem[Blacoe \& Lapata(2012)Blacoe and Lapata]{Blacoe2012A}
Blacoe, William and Lapata, Mirella.
\newblock A comparison of vector-based representations for semantic
  composition.
\newblock In \emph{Joint Conference on Empirical Methods in Natural Language
  Processing and Computational Natural Language Learning}, pp.\  546--556,
  2012.

\bibitem[Chitra \& Rajkumar(2016)Chitra and Rajkumar]{Chitra2016Plagiarism}
Chitra, A. and Rajkumar, Anupriya.
\newblock Plagiarism detection using machine learning-based paraphrase
  recognizer.
\newblock \emph{Journal of Intelligent Systems}, 25\penalty0 (3):\penalty0
  351--359, 2016.

\bibitem[Fernando et~al.(2017)Fernando, Banarse, Blundell, Zwols, Ha, Rusu,
  Pritzel, and Wierstra]{Fernando2017PathNet}
Fernando, Chrisantha, Banarse, Dylan, Blundell, Charles, Zwols, Yori, Ha,
  David, Rusu, Andrei~A, Pritzel, Alexander, and Wierstra, Daan.
\newblock Pathnet: Evolution channels gradient descent in super neural
  networks.
\newblock \emph{Arxiv}, 2017.

\bibitem[Graves et~al.(2014)Graves, Wayne, and Danihelka]{Graves2014Neural}
Graves, Alex, Wayne, Greg, and Danihelka, Ivo.
\newblock Neural turing machines.
\newblock \emph{Computer Science}, 2014.

\bibitem[Gulcehre et~al.(2016{\natexlab{a}})Gulcehre, Ahn, Nallapati, Zhou, and
  Bengio]{Gulcehre2016Pointing}
Gulcehre, Caglar, Ahn, Sungjin, Nallapati, Ramesh, Zhou, Bowen, and Bengio,
  Yoshua.
\newblock Pointing the unknown words.
\newblock In \emph{Proceedings of the 54th Annual Meeting of the Association
  for Computational Linguistics}, pp.\  140--149, 2016{\natexlab{a}}.

\bibitem[Gulcehre et~al.(2016{\natexlab{b}})Gulcehre, Chandar, Cho, and
  Bengio]{Gulcehre2016Dynamic}
Gulcehre, Caglar, Chandar, Sarath, Cho, Kyunghyun, and Bengio, Yoshua.
\newblock Dynamic neural turing machine with soft and hard addressing schemes.
\newblock \emph{Arxiv}, 2016{\natexlab{b}}.

\bibitem[Heilman \& Smith(2010)Heilman and Smith]{Heilman2010Tree}
Heilman, Michael and Smith, Noah~A.
\newblock Tree edit models for recognizing textual entailments, paraphrases,
  and answers to questions.
\newblock In \emph{Human Language Technologies: Conference of the North
  American Chapter of the Association of Computational Linguistics,
  Proceedings, June 2-4, 2010, Los Angeles, California, USA}, pp.\  1011--1019,
  2010.

\bibitem[Hochreiter \& Schmidhuber(1997)Hochreiter and
  Schmidhuber]{Hochreiter1997Long}
Hochreiter, Sepp and Schmidhuber, Jürgen.
\newblock Long short-term memory.
\newblock \emph{Neural Computation}, 9\penalty0 (8):\penalty0 1735, 1997.

\bibitem[Kravchenko(2017)]{Kravchenko2017Paraphrase}
Kravchenko, Dmitry.
\newblock Paraphrase detection using machine translation and textual similarity
  algorithms.
\newblock In \emph{Conference on Artificial Intelligence and Natural Language},
  pp.\  277--292, 2017.

\bibitem[Lai \& Hockenmaier(2014)Lai and Hockenmaier]{Lai2014Illinois}
Lai, Alice and Hockenmaier, Julia.
\newblock Illinois-lh: A denotational and distributional approach to semantics.
\newblock In \emph{International Workshop on Semantic Evaluation}, pp.\
  329--334, 2014.

\bibitem[Pennington et~al.(2014)Pennington, Socher, and
  Manning]{Pennington2014Glove}
Pennington, Jeffrey, Socher, Richard, and Manning, Christopher.
\newblock Glove: Global vectors for word representation.
\newblock In \emph{Conference on Empirical Methods in Natural Language
  Processing}, pp.\  1532--1543, 2014.

\bibitem[Pham et~al.(2018)Pham, Guan, Zoph, Le, and Dean]{Pham2018Efficient}
Pham, Hieu, Guan, Melody~Y, Zoph, Barret, Le, Quoc~V, and Dean, Jeff.
\newblock Efficient neural architecture search via parameter sharing.
\newblock \emph{Arxiv}, 2018.

\bibitem[Shen \& Lapata(2007)Shen and Lapata]{Shen2007Using}
Shen, Dan and Lapata, Mirella.
\newblock Using semantic roles to improve question answering.
\newblock In \emph{EMNLP-CoNLL 2007, Proceedings of the 2007 Joint Conference
  on Empirical Methods in Natural Language Processing and Computational Natural
  Language Learning, June 28-30, 2007, Prague, Czech Republic}, pp.\  12--21,
  2007.

\bibitem[Socher et~al.(2011)Socher, Huang, Pennin, Manning, and
  Ng]{socher2011dynamic}
Socher, Richard, Huang, Eric~H, Pennin, Jeffrey, Manning, Christopher~D, and
  Ng, Andrew~Y.
\newblock Dynamic pooling and unfolding recursive autoencoders for paraphrase
  detection.
\newblock In \emph{Advances in Neural Information Processing Systems}, pp.\
  801--809, 2011.

\bibitem[Wang et~al.(2009)Wang, Smith, and Mitamura]{Wang2009What}
Wang, Mengqiu, Smith, Noah~A., and Mitamura, Teruko.
\newblock What is the jeopardy model? a quasi-synchronous grammar for qa.
\newblock In \emph{EMNLP-CoNLL 2007, Proceedings of the 2007 Joint Conference
  on Empirical Methods in Natural Language Processing and Computational Natural
  Language Learning, June 28-30, 2007, Prague, Czech Republic}, pp.\  22--32,
  2009.

\bibitem[Wang et~al.(2016{\natexlab{a}})Wang, Mi, and
  Ittycheriah]{Wang2016Semi}
Wang, Zhiguo, Mi, Haitao, and Ittycheriah, Abraham.
\newblock Semi-supervised clustering for short text via deep representation
  learning.
\newblock In \emph{the 20th SIGNLL Conference on Computational Natural Language
  Learning (CoNLL)}, 2016{\natexlab{a}}.

\bibitem[Wang et~al.(2016{\natexlab{b}})Wang, Mi, and
  Ittycheriah]{Wang2016Sentence}
Wang, Zhiguo, Mi, Haitao, and Ittycheriah, Abraham.
\newblock Sentence similarity learning by lexical decomposition and
  composition.
\newblock In \emph{the 26th International Conference on Computational
  Linguistics}, 2016{\natexlab{b}}.

\bibitem[Wang et~al.(2017)Wang, Hamza, and Florian]{Wang2017Bilateral}
Wang, Zhiguo, Hamza, Wael, and Florian, Radu.
\newblock Bilateral multi-perspective matching for natural language sentences.
\newblock In \emph{Twenty-Sixth International Joint Conference on Artificial
  Intelligence (IJCAI-17)}, 2017.

\bibitem[Wright(1990)]{Wright1990Speeding}
Wright, M.~B.
\newblock Speeding up the hungarian algorithm.
\newblock \emph{Computers and Operations Research}, 17\penalty0 (1):\penalty0
  95--96, 1990.

\bibitem[Yang et~al.(2015)Yang, Yih, and Meek]{Yang2015WikiQA}
Yang, Yi, Yih, Wen-tau, and Meek, Christopher.
\newblock Wikiqa: A challenge dataset for open-domain question answering.
\newblock In \emph{Conference on Empirical Methods on Natural Language
  Processing}, pp.\  2013--2018, 2015.

\bibitem[Yao et~al.(2013)Yao, Durme, Callison-Burch, and Clark]{Yao2013Semi}
Yao, X., Durme, B.~Van, Callison-Burch, C., and Clark, P.
\newblock Semi-markov phrase-based monolingual alignment.
\newblock \emph{2013 Conference on Empirical Methods in Natural Language
  Processing}, 2013.

\bibitem[Yin et~al.(2015)Yin, Schütze, Xiang, and Zhou]{Yin2015ABCNN}
Yin, Wenpeng, Schütze, Hinrich, Xiang, Bing, and Zhou, Bowen.
\newblock Abcnn: Attention-based convolutional neural network for modeling
  sentence pairs.
\newblock \emph{Computer Science}, 2015.

\bibitem[Yu et~al.(2014)Yu, Hermann, Blunsom, and Pulman]{yu2014deep}
Yu, Lei, Hermann, Karl~Moritz, Blunsom, Phil, and Pulman, Stephen.
\newblock Deep learning for answer sentence selection.
\newblock \emph{arXiv preprint arXiv:1412.1632}, 2014.

\bibitem[Zeiler(2012)]{Zeiler2012ADADELTA}
Zeiler, Matthew~D.
\newblock Adadelta: An adaptive learning rate method.
\newblock \emph{Computer Science}, 2012.

\end{thebibliography}
\bibliographystyle{plain}

\end{document}